\pdfoutput=1
\documentclass[11pt]{article}

\usepackage{acl}

\usepackage{times}
\usepackage{latexsym}
\usepackage{booktabs} % To thicken table lines

\usepackage[T1]{fontenc}
\usepackage[utf8]{inputenc}

\usepackage{microtype}

\usepackage{amsmath,amssymb}
\usepackage{color,soul}
\usepackage{enumitem} 
\usepackage{graphicx}
\usepackage{multirow}
\usepackage{tikz}
\usepackage{tcolorbox}
\usepackage{svg}

\newcommand{\indep}{\perp\!\!\!\!\perp}

\title{Quantifying the Task-Specific Information in Text-Based Classifications}

\author{Zining Zhu$^{1,2}$, Aparna Balagopalan$^{1,2}$, Marzyeh Ghassemi$^{3}$, Frank Rudzicz$^{1,2,4}$\\
$^1$University of Toronto $^2$Vector Institute for Artificial Intelligence $^3$MIT $^4$Unity Health Toronto \\
\texttt{\{zining, aparna\}@cs.toronto.edu}\\ \texttt{mghassem@mit.edu, frank@cs.toronto.edu}}

\begin{document}
\maketitle
\begin{abstract}
Recently, neural natural language models have attained state-of-the-art performance on a wide variety of tasks, but the high performance can result from superficial, surface-level cues~\cite{Bender2020,Niven2020}. These surface cues, as the ``shortcuts'' inherent in the datasets, do not contribute to the \textit{task-specific information} (TSI) of the classification tasks. While it is essential to look at the model performance, it is also important to understand the datasets. In this paper, we consider this question: Apart from the information introduced by the shortcut features, how much task-specific information is required to classify a dataset? We formulate this quantity in an information-theoretic framework. While this quantity is hard to compute, we approximate it with a fast and stable method. TSI quantifies the amount of linguistic knowledge modulo a set of predefined shortcuts -- that contributes to classifying a sample from each dataset. This framework allows us to compare across datasets, saying that, apart from a set of ``shortcut features'', classifying each sample in the Multi-NLI task involves around 0.4 nats more TSI than in the Quora Question Pair.
\end{abstract}

\section{Introduction}
Neural natural language processing (NLP) models have attained state-of-the-art classification tasks, including natural language inference, sentiment analysis, and textual similarity~\cite{devlin2019bert,yang2019xlnet}. 
What drives this performance? A popular argument is: neural models learn certain linguistic skills for these tasks, and their representations encode linguistic knowledge~\cite{lakretz-etal-2019-emergence,hewitt-manning-2019-structural-probe,Chen2019DiscoEval,tenney-etal-2019-bert,jiang2019evaluating,Zhu2020RSTprobe,ettinger2020bert}.
How can neural models encode this linguistic knowledge? \citet{Alain2017} suggested that, by attending to datasets, neural NLP models gradually learn to preserve useful, task-specific information while discarding the rest. In this way, the task-specific information is ``distilled'' in the neural network models. 
There are many text-based classification tasks (e.g., \citet{Multi-NLI}), each of which requires some amount of linguistic information to classify that the neural networks distill along the way.

\begin{figure}[t]
    \centering
    \includegraphics[width=.8\linewidth]{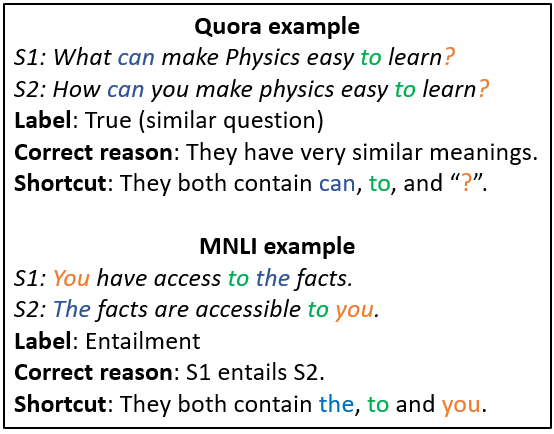}
    \caption{Classifiers can rely on ``shortcut features'' to reach the correct predictions, but this strategy is not generalizable, since the classifiers do not learn the real linguistic knowledge. Shortcut features, including the occurrence of punctuation marks (e.g., ``{\color{orange}?}'') and stopwords (e.g., {\color{blue}can, the}, {\color{green}to}, {\color{orange}you}), are prevalent in datasets, but should not be part of the linguistic knowledge required to classify. % Updated the example. "easy to learn" is actually a borderline one -- we can argue that it refers to part of the true semantics.
    We propose a method to quantify how much task-specific, shortcut-irrelevant information remains in the datasets.}
    \label{fig:shortcut_example}
\end{figure}

The inquiry into the information regime of models leads to an appealing goal in explainable AI~\cite{doran2018does}: to infer the amount of task-specific, linguistic knowledge required for a given task in information-theoretic terms. With this unified metric, we will be able to compare across text-based classification tasks. Typically, classification accuracy and loss are used for comparison. However, recent research showed that a low cross-entropy loss might result from the information that is correlative but not causative to the prediction tasks. This is the ``shortcut learning'' problem, and it happens in a wide variety of classification tasks~\cite{ThomasMcCoy2019,Geirhos2020shortcut,Niven2020,Misra2020,Stali2020} -- even in human cognition, where study participants figure out more accessible ways to solve testing tasks~\cite{geirhos2020unintended}. 

Figure \ref{fig:shortcut_example} presents two examples of shortcuts, where we could make predictions based on shortcuts that are irrelevant to the linguistic knowledge of the tasks. 
Therefore, shortcuts constitute a gap between how much \textit{is learned} and how much \textit{should be learned} to classify the task. Following the motivations of recent causal analysis papers (e.g., \citet{elazar_amnesic_2021,pryzant-etal-2021-causal}), we want to factor out the impact of the shortcuts while still quantifying the amount of information a neural network model needs to learn for a task.

This paper presents a framework to separate the surface-level shortcuts from the deeper information. We quantify the ``task-specific information'' (TSI) that is not part of the spurious correlations. TSI is hard to compute numerically, but we use a method based on a Bayesian formulation to approximate this quantity (\S \ref{sec:task-specific-information}). The computation only requires computing cross-entropy losses on a pair of classification tasks.
We discuss the proper choice of configurations to compute the TSI (Secs. \ref{subsec:model-consistency},\ref{subsec:experiments:Xb-choices}). 
Our method is stable across dataset sizes (\S~\ref{subsection:experiments:dataset-size}), and is easier to compute than existing entropy estimators (\S~\ref{subsec:alternative-estimators}).

Overall, the TSI framework quantifies the ``linguistic knowledge'' required to perform text-based classifications and further allows principled comparisons of the degrees of linguistic knowledge across a wide range of classification tasks. For example, the classification task in MNLI dataset~\cite{Multi-NLI} requires about 0.25 nats more TSI than the sentiment detection task with IMDB movie reviews~\cite{Maas2011IMDB}, and around 0.4 nats more than the textual similarity detection task with the QQP dataset~\cite{Wang2019} (\S~\ref{subsec:full-dataset}), given a fixed set of shortcuts.

\section{Related Work}
Our work is related to prior work in identifying and isolating spurious artifacts (``shortcuts'') in text-based prediction tasks, probing language embeddings for various linguistic phenomena, and analyzing dataset statistics.
\paragraph{Shortcut learning}
Deep neural networks can overtly rely on superficial heuristics, which allows them to perform well on standard benchmarks but prohibits generalization to real-world scenarios. 
\citet{Geirhos2020shortcut} called this problem ``shortcut learning'' and referred to these heuristics as ``shortcuts". On text-based classification datasets, shortcuts appear in the form of spurious statistical cues. These include the warrants for argument reasoning~\cite{Niven2020}, syntax heuristics and lexical overlaps in natural language inference~\cite{ThomasMcCoy2019}, and relevant words (``semantic priming'')~\cite{Misra2020}. These spurious surface cues do not contribute to task-specific information.

By carefully constructing test sets that do not have these statistical cues and spurious associations, such shortcuts can be diagnosed~\cite{glockner-etal-2018-breaking,gardner2020evaluating}. 
\citet{kaushik2019learning} counterfactually augmented text snippets in several sentiment-classification datasets via crowd-sourcing by applying minimal changes to the original text to flip the prediction label. \citet{Rosenman2020} used challenge sets to reveal the ``learning by heuristics'' problem in the relation extraction task. In contrast to our work, none of these prior works formulate the issue of shortcut learning using information theory. Another strategy to factor out known dataset biases is debiasing algorithms, such as the residual fitting algorithm \citep{He2019}.

\paragraph{Probing}
The probing literature inspires our approach to analyzing the information in neural language models. According to \citet{Alain2017}, the task of probing asks, ``is there any information about factor \underline{\hspace{2em}} in this part of the model?'' Following this line, many subsequent papers queried the amount of knowledge from various parts of neural models. These included  syntax-related~\cite{lakretz-etal-2019-emergence,hewitt-manning-2019-structural-probe}, semantic-related~\cite{tenney-etal-2019-bert}, and discourse-related information~\cite{Chen2019DiscoEval,koto-etal-2021-discourse}.
Towards developing reliable probing methods, several papers proposed control mechanisms~\cite{pimentel2020information,zhu-rudzicz-2020-information}. With a collection of imperfect classifiers, we can combine to adjust for potential confounds. Our analyses are motivated by this idea, but we study the classification instead of the probing regime.

\paragraph{Understanding the datasets}
In machine learning and NLP literature, several works studied the ``difficulty'' of datasets \citep{blache2011predicting, Gupta2014, Collins2018, Jain2020DataQuality}, but they did not consider factoring out the impacts of shortcuts.
\citet{Damour2020} framed the shortcut learning issue as an underspecification problem: There is not enough information in training set to distinguish between spurious artifacts and the inductive biases (or rather, the linguistic knowledge).
Recently, researchers have analyzed the behavior of models on individual samples during training to diagnose datasets~\cite{tu2020empirical,kumar2019topics}. \citet{han-etal-2020-explaining} used influence functions to identify influential training samples and characterize the artifacts in datasets. \citet{Swayamdipta2020} computed metrics of training dynamics of a model, i.e., the prediction confidence and variability, to map a ``cartography'' of the data samples. 
\citet{warstadt2020learning} introduced a dataset to study linguistic feature learning versus generalization in the RoBERTa base model and considered a probing setup with a control task to investigate the inductive biases of a pretrained model at the fine-tuning time. \citet{lovering_predicting_2021} found that the extent that a feature influences a model's decisions is affected by the probing extractability and its co-occurrence rate with the label.
These works have a common intuition: we should study the datasets to study the spurious correlation (shortcuts). We follow this line of research and quantify the information of shortcuts in the datasets. 

\paragraph{Mutual information}
Our work is related to information theory formulations about machine learning. \citet{voita-titov-2020-information} proposed two approaches to measure the minimum description lengths of probing. \citet{li-eisner-2019-specializing} used a method based on variational information bottleneck to compress word embeddings and improve parser performances. \citet{steinke20ConditionalMI} proposed a formulation of conditional mutual information that can be used to reason about the generalization properties of machine learning models.  
Empirically, our proposed method (using the difference of a pair of cross-entropy losses) echoes what \citet{xu2020theory} defined as the ``predictive $\mathcal{V}$-information''. We derive TSI from a different perspective from the $\mathcal{V}$-information. We elaborate in \S \ref{sec:task-specific-information}. There are several contemporaneous works. \citet{o2021context} uses $\mathcal{V}$-information to study the effects of each context feature independently. \citet{ethayarajh-2021-information} uses pointwise $\mathcal{V}$-information to describe the dataset difficulty. 

\begin{figure}[t]
    \centering
    \includegraphics[width=.3\linewidth]{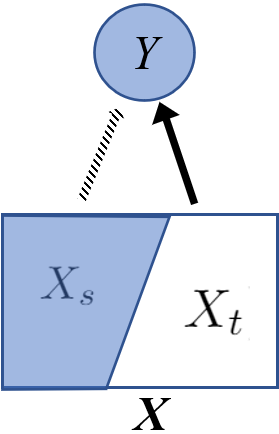}
    \caption{An illustration of the relationships between the text data $X$, containing a shortcut part $X_s$, and an unmeasurable task-specific part $X_t$, as well as the task label $Y$. The solid arrow indicates a causal relationship, while the dashed arrow indicates a spurious correlation. We want to factor out the observable $X_s$ from this graph.}
    \label{fig:causal-model}
\end{figure}

\section{Learning Task-Specific Information}
\label{sec:task-specific-information}

This section presents our framework to quantify the task-specific information. 

\subsection{Removing the shortcuts}
Consider a dataset of data points $\{(x_i, y_i)\}_{n=1}^{N}$, where $x_i\in \mathbb{R}^m$ is the feature vector, and $y_i$ is the label. Let the random variable $X$ represent all possible input features, and the random variable $Y$ represent the task labels.

In our framework, the input random variable $X$ constitutes of the shortcut part, denoted by a random variable $X_s$, and the task-specific part, an un-measurable $X_t$. In other words, $X=f(X_s, X_t)$, where $X_s \indep X_t$, and $f(\cdot)$ can be any composition function. Their dependency relationships can be described by Figure \ref{fig:causal-model}. This allows us to write the distributions as:
\begin{align}
    p(Y\,|\,X) = p(Y\,|\,X_t)p(Y\,|\,X_s)\underbrace{\frac{p(X_t)p(X_s)}{p(X)p(Y)}}_{\text{prior}}
\label{eq:bayesian_Xt}
\end{align}
When $X_s \indep X_t$, $p(X)=p(X_t)p(X_s)$, so the prior term degenerates into $\frac{1}{p(Y)}$. Now, the mutual information between $Y$ and $X_t$ is:

\begin{align}
\begin{split}
I(Y;X_t)&=\mathbb{E} \text{ log }\frac{p(Y,X_t)}{p(X_t)p(Y)} = \mathbb{E} \text{ log }\frac{p(Y\,|\,X_t)}{p(Y)} \\
&= \mathbb{E} \text{ log } \frac{1}{p(Y\,|\,X_s)} - \mathbb{E} \text{ log } \frac{1}{p(Y\,|\,X)}\\
&= H(Y\,|\,X_s) - H(Y\,|\,X)
\label{eq:I_eq_diff_entropy}
\end{split}
\end{align}
where the expectations are taken over the distribution implicitly defined by the data $\{x_i, y_i\}_{i=1}^N$. The equation in the second last line is acquired by substituting in Eq. \ref{eq:bayesian_Xt}.

\subsection{Interpreting the model performance}
Empirically, a model learning this task (e.g., a BERT \citep{devlin2019bert} with a fully connected layer on top) approximates the true, unknown distribution $p(Y\,|\,X)$. Let $q(Y\,|\,X)$ describe the learned model, then by definition:
\begin{align}
    H(Y\,|\,X) = \text{NLL}_{Y\,|\,X} - \text{KL}(p\,\|\,q)
\end{align}
where NLL denotes the negative log likelihood loss,\footnote{We assume continuous distributions, so $\text{NLL}_{Y\,|\,X}=\Sigma_{x\in \text{data}} -\text{log}q(Y\,|\,X)$ equals the cross entropy $\mathbb{E}_{x} -\text{log}q(Y\,|\,X)$.} KL is the Kullback-Leibler divergence, $p$ and $q$ are the short-hand notations of $p(Y\,|\,X)$ and $q(Y\,|\,X)$ respectively, and 
\begin{align}
    \text{NLL}_{Y\,|\,X}=\mathbb{E}_{p(X)} \text{ log }\frac{1}{q(Y\,|\,X)}
\end{align} is the cross-entropy loss. In this paper, we will use $\text{NLL}$ to refer to the cross-entropy loss, for clarity.

A well-trained model would have high performance: a high accuracy, a low $\text{KL}(p\,\|\,q)$ divergence, and a low cross-entropy loss. However, as mentioned before, this could result from the model ``taking shortcuts'', predicting the task labels $Y$ from the shortcuts $X_s$.

\subsection{Computing TSI needs a control task}
Here we consider a control task to specify the features that might benefit the classification but do not contribute to the linguistic knowledge required for the models to perform the task correctly. Figure \ref{fig:shortcut_example} describes some shortcuts. We include the details in the Experiment below.

We refer to the classifier trained only on the shortcuts as the control model. When trained, the control model approximates the unknown distribution $p(Y\,|\,X_s)$ with an empirical distribution, $q(Y\,|\,X_s)$. 

\paragraph{Definition 1:} The \textit{task-specific information} (TSI) in the classification task (described by $X,Y$) with respect to the shortcut $X_s$ is quantified by:
\begin{align}
\begin{split}
&I(Y;X_t) = \underbrace{\text{NLL}_{Y\,|\,X_s} - \text{NLL}_{Y|X}}_{\textrm{Known}} + \\
&\underbrace{\text{KL}(p_{Y\,|\,X}\,\|\,q_{Y\,|\,X}) - \text{KL}(p_{Y\,|\,X_s}\,\|\,q_{Y\,|\,X_s})}_{\textrm{Unknown}}
\label{eq:info_y_xg}
\end{split}
\end{align}

Similarly, $\text{NLL}_{Y\,|\,X_s}$ is the cross-entropy loss of the control task. They can be measured empirically, so we mark them as ``known''.

\subsection{On the scales of the intractable KLs}
In Eq. \ref{eq:info_y_xg}, the two ``known'' terms constitute of the predictive $\mathcal{V}$-information \citep{xu2020theory} from $X_t$ to $Y$. Additionally, $I(Y;X_t)$ contains two intractable KL terms. As a sanity check, we use a collection of synthetic datasets to estimate their scales. Following are the distributions to generate these toy datasets $\{X,Y\}$:
\begin{align*}
    &X_j \sim \text{Bernoulli}(p_x) \text{, where }j\in \{1,2,.., m\} \\
    &X = [X_1, X_2, ..., X_m] \\
    &Y = g(X_1, ..., X_m) + \epsilon \text{, where }\epsilon\sim \text{Bernoulli}(p_y) 
\end{align*}
where $m$ specifies the number of input features, and $g(X_1, ..., X_m)$ is a deterministic function. This construction allows an exact computation of the conditional entropy $H(Y\ |\ X)$. On the other hand, we compute the cross-entropy $\text{NLL}_{Y\ |\ X}$ by training a default scikit-learn MLPClassifier $q(Y\ |\ X)$ on the train portion of $\{X, Y\}$. Then, the difference between the dev loss and the conditional entropy is the KL values resulting from the imperfect classifier.

\begin{figure}[t]
    \centering
    \includesvg[width=\linewidth]{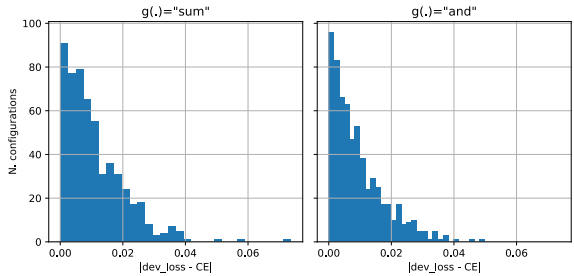}
    \caption{The histograms of $|NLL-H(Y\ |\ X)|$, i.e., the estimated scales of $\text{KL}(p\ \|\ q)$, with the \texttt{sum} and \texttt{and} option respectively.}
    \label{fig:KL_bounds}
\end{figure}

We generate toy datasets with different values of $m$ ($2\leq m \leq 10$), $p_x$ and $p_y$ (between 0.1 and 0.9). For $g(\cdot)$, we use two options: 
\begin{itemize}[nosep]
    \item \texttt{sum}: $g(X)=\sum_j X_j$
    \item \texttt{and}: $g(X)=X_1 \wedge X_2 \wedge ... \wedge X_m$
\end{itemize}
Figure \ref{fig:KL_bounds} show the histograms of the two options, respectively. In 99.5\% (1184 of 1190) configurations, the dev losses are within 0.04 nats away from $H(Y\ |\ X)$. In other words, the scales of the $KL(p\ \|\ q)$ are estimated to be one magnitude smaller than those of $I(Y;X_t)$. Additionally, a recent paper, \citet{pimentel2021bayesian} shows that the difference of a pair of cross entropy (they call it Bayesian Mutual Information) converges to the mutual information when there are infinite number of data points. In the subsequent analysis, we empirically ignore the intractable KL terms.

\subsection{Understanding TSI}
\label{subsec:understanding-info}
Before moving to the computation, let us first briefly discuss some properties of TSI.

\textit{Lower bound.} $\text{TSI}\geq 0$, where equality is reached when the information from the shortcuts (e.g., the presence of specific tokens) is sufficient for classification, so the model does not have to learn any task-specific knowledge to perform perfectly.

\textit{Upper bound.} $\text{TSI}\leq H(Y)$, where the equality is reached when $H(Y\,|\,X_t)=0$, i.e., the task label $Y$ is a deterministic function of the task-specific variable $X_t$.
Further, for a task with $m$ distinct labels, Jensen-Shannon inequality gives us $H(Y)\leq \text{log }m$ nats.\footnote{Throughout this paper, we use nats (instead of bits) as the unit for measuring the information-theoretic terms.} When $m=2$ and $3$, the TSI would be correspondingly upper-bounded by $\text{log}2\approx 0.693$ and $\text{log}3\approx 1.097$, respectively. When the number of classes $m$ increases, the upper bound of TSI increases, resembling what \citet{Gupta2014} mentioned about how a larger number of classes contribute to the increased cross-entropy.

\textit{An on-average metric.} TSI is averaged across the dataset samples, allowing comparison across datasets with different sizes. We can compare the TSI scores of a dataset with 50,000 samples (e.g., IMDB \citep{Maas2011IMDB}) to that of a dataset with 400,000 samples (e.g., Quora Question Pairs) to directly compare their ``linguistic informativeness''. 
We discuss further about the dataset sizes in \S \ref{subsection:experiments:dataset-size}.

\textit{Quantity but not form.} TSI quantifies the amount rather than describes the actual type of information required to classify a task. The former computes an aggregate metric, while the latter requires a deep understanding of the task knowledge. This paper considers the former.

\section{Experiments}
\label{sec:experiments}

\subsection{Datasets} 
We run experiments on several popular benchmarking datasets (in English) that test various linguistic abilities, including sentiment and attitude detection (Yelp and IMDB), entailment recognition (MNLI), and semantic similarity understanding (QQP). The dataset details are in Appendix \ref{sec:dataset-details}.

\subsection{Control task features} The features for the control task need to be scalars. In the experiments, we use the following features to illustrate the application of our framework.

\paragraph{The occurrences of punctuations} We count the punctuation in each input text sample and normalize by the number of tokens in the sentence. If a sample constitutes a pair of sentences, we concatenate the two sentences. Following is an example.
\begin{tcolorbox}
You have access to the facts \textul{.} The facts are accessible to you \textul{.}
\end{tcolorbox}
There are $N=2$ occurrences of punctuations in the (concatenated) sentence with length $L=14$, so the ``occurrence of punctuation'' feature is $\frac{2}{14}$.

\paragraph{The occurrence of stopwords} We count the stopwords (modulo the negation words including ``no'', ``nor'', ``don't'' and ``weren't'') and normalize by the token length of the example. We concatenate the two sentences for the samples consisting of a pair of sentences similar to the punctuation feature. Following is an example.
\begin{tcolorbox}
\textul{You} \textul{have} access \textul{to} \textul{the} facts . \textul{The} facts \textul{are} accessible \textul{to} \textul{you} .
\end{tcolorbox}
There are $N=8$ occurrences of stopwords in this sentence with length $L=14$, so the ``occurrence of stopword'' feature is $\frac{8}{14}$. Note that some stopwords do have semantic roles. For example, \texttt{I}, \texttt{you} and \texttt{they} can specify the person(s) in the situations. Additionally, one could argue that the choice of stopwords between, e.g., \texttt{I} and \texttt{me} could indicate the role of the speaker, and so on. Therefore one could argue that the occurrence of stopwords can be a non-shortcut, dependent on the actual task. However, one can also argue for the opposite, since the information provided by these semantic roles seems irrelevant to various classification tasks -- for example, both ``I like this movie'' and ``You like this movie'' would indicate a positive movie review. This collection serves as an example that the TSI framework allows considering a collection of semantically nontrivial words.

\paragraph{The overlapping of paired sentences} For each pair of sentence $(s_1, s_2)$, we use the number of overlapped tokens (relative to each of the two sentence lengths) to describe the extent of lexical overlapping. Following is an example.
\begin{tcolorbox}
\begin{itemize}[nosep]
    \item What \textul{can} \textul{make} \textul{Physics} \textul{easy} \textul{to} \textul{learn} \textul{?}
    \item How \textul{can} you \textul{make} \textul{Physics} \textul{easy} \textul{to} \textul{learn} \textul{?}
\end{itemize}
\end{tcolorbox}
The two ``lexical overlap'' features for this sentence pair are overlap\_1=$\frac{8}{9}$, overlap\_2=$\frac{8}{10}$.

\subsection{Classification models} For training $q(Y\,|\,X)$ models, we use BERT \citep{devlin2019bert}, RoBERTa \citep{Liu2019roberta}, and ALBERT \citep{Lan2020ALBERT}, all on the base configuration (12 layers), with a fully connected head. Such transformer-based configurations are the state-of-the-art on classification tasks. We adopt the configurations of \citep{devlin2019bert}: we concatenate the input sentences (for MNLI and QQP) and take the \texttt{[CLS]} token representations to pass in the fully connected head. The training hyperparameters follow the configurations recommended in the literature \citep{devlin2019bert,Liu2019roberta,Lan2020ALBERT,wolf2019huggingface}. For training $q(Y\,|\,X_s)$ models, we use MLPClassifier from scikit-learn \citep{scikit-learn}. We list the details in Appendix \ref{sec:hyperparameters}. 

\begin{figure}[t]
    \centering
    \includesvg[width=\linewidth]{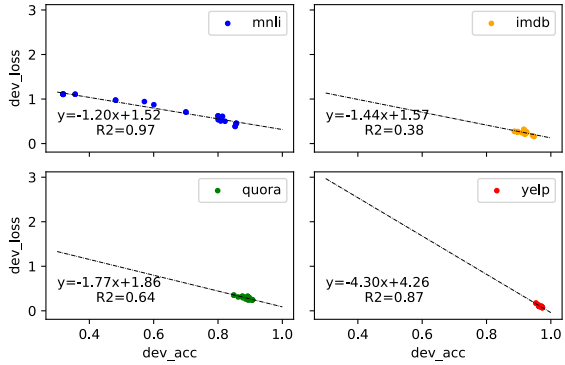}
    \caption{A scatter plot of the accuracy against dev loss of models trained on full datasets.}
    \label{fig:loss_acc_corr}
\end{figure}
\begin{figure}[t]
    \centering
    \includesvg[width=\linewidth]{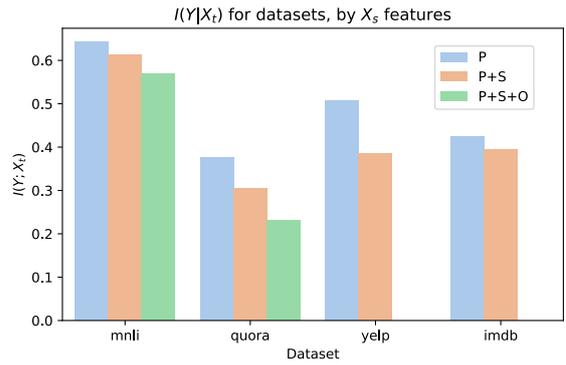}
    \caption{Estimates of TSI with different choices of shortcut features and the best models. Note that the \texttt{O} (lexical overlapping) heuristics only apply for MNLI and QQP, while the \texttt{P} (punctuation) and \texttt{S} (stopwords) heuristics apply to all four tasks. For each task, as more features are excluded, we can see the estimate decreases. Unless specifically mentioned, we consider TSI$^{\text{P+S}}$ for all tasks henceforth.}
    \label{fig:xb_variations}
\end{figure}

% Moved forward for formatting
\begin{figure*}
    \centering
    \includesvg[width=.8\linewidth]{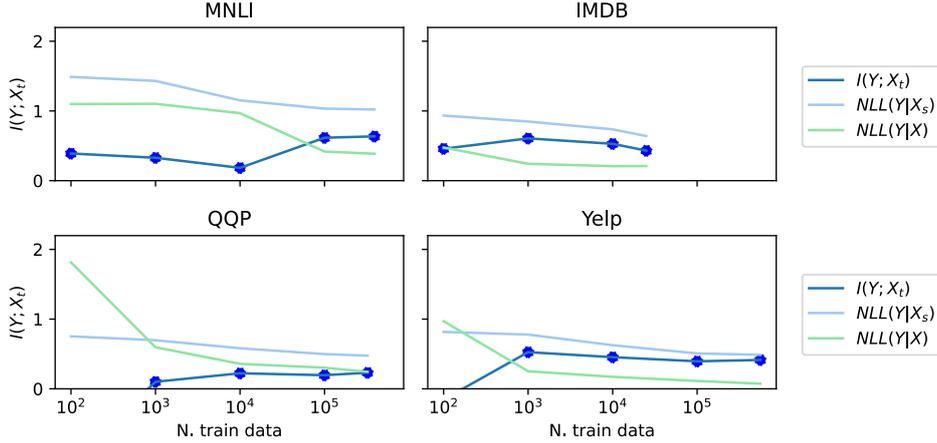}
    \caption{The $I(Y;X_t)$ estimation when we subsample different sizes of datasets.}
    \label{fig:subset_impacts}
\end{figure*}

\section{Discussions}
\subsection{Estimating TSI with an suboptimal model}
\label{subsec:model-consistency}
Each $\{X, X_s, Y\}$ configuration uniquely determines the $I(Y;X_t)$ value. 
Ideally, the models that perfectly fit the dataset distributions $p(Y|X)$ and $p(Y\ |\ X_s)$ can precisely estimate $I(Y\ ;\ X_t)$. Among all empirical models, the highest performing models approximate $I(Y;X_t)$ the most closely, since they lead to KL values (of Eq. \ref{eq:info_y_xg}) that are the smallest. Therefore, we report the results from finetuning the best of BERT, RoBERTa, and ALBERT, and we recommend using the best possible model.

Empirically, the model at hand might have an accuracy of several points lower than the top model at the GLUE leaderboard. How far do the entropy values of the imperfect models differ from those of the SOTA models (which usually only the accuracies are available)? Figure \ref{fig:loss_acc_corr} plots the correlations between the cross-entropy losses and the accuracies of the non-degenerative finetuned $q(Y\,|\,X)$ models. Interestingly, except for IMDB, the results show linear trends, with the slopes and intercepts varying from task to task. The slopes of the trendlines could be used to interpolate the validation losses of the suboptimal models.

\subsection{TSI and the choice of shortcuts}
\label{subsec:experiments:Xb-choices}
To enable cross-task comparisons, our framework considers TSI with respect to the fixed set of shortcuts. For example, apart from lexical overlap, how much linguistic information is there in the classification? The choice of shortcut features affects the cross-entropy losses, hence the TSI.

Figure~\ref{fig:xb_variations} reports the TSI estimations with various choices of shortcut features (additional results are in the Appendix). As we add features to the $X_s$ set, $\text{NLL}_{Y\,|\,X_s}$ decreases, leading to a corresponding decrease in TSI. The lexical overlap feature exacerbates this decrease to $X_s$ for MNLI. This follows our intuition since the syntactic heuristics such as lexical overlap have been identified as fallible heuristics for MNLI in prior work~\cite{ThomasMcCoy2019}, and though lexemes are shortcut features, they do encode semantics.

\textit{On the completeness of shortcuts}. We do not aim at the unrealistic goal of exhausting all possible shortcuts. Instead, we present a framework where the contribution of the shortcuts, once identified, can be factored out. The TSI framework can generalize to additional shortcuts.

\textit{Generalization of features}. We identified some features as ``shortcut features''. Dependent on the goal of analysis, one can apply other features (e.g., the length of sentences). In addition, automatic identification of shortcut features $X_s$ method (e.g., approaches similar to those of \citet{wang-culotta-2020-identifying}) may be used as well.

\subsection{How stable is TSI to dataset size?}
\label{subsection:experiments:dataset-size}
To evaluate the effects of dataset size, we reduce the training sets with stratified sampling while assessing on the same validation set.
As shown in Figure \ref{fig:subset_impacts}, the robustness of TSI estimations regarding the subset size differs across datasets. For MNLI, the estimation started to fluctuate starting from 25\% of the original size. However, the estimates for IMDB, Quora, and Yelp remain relatively stable until we reduce the train set sizes to as little as $\sim5\%$.

For both the $Y|X$ and $Y|X_s$ classification, the minimum reachable cross-entropy losses increase as we reduce the dataset sizes. A possible reason is that downsampling changes the data distribution and leads to mismatches between the train and the validation distributions. Similar effects are described in e.g., \citet{gardner2020evaluating}. Note that as we reduce the dataset sizes, $H(Y|X)$ rises faster than $H(Y|X_s)$, indicating that the deeper, task-specific knowledge requires more data to capture than those shortcut knowledge, echoing the finding of \citet{warstadt2020learning}.

\subsection{What about alternative estimators?}
\label{subsec:alternative-estimators}
Previous works have proposed several mutual information estimators based on setting up optimization goals, e.g., BA \citep{agakov_im_2004}, DV \citep{donsker1975asymptotic}, NWJ \citep{nguyen2010estimating}, MINE \citep{belghazi2018mutual}, CPC \citep{oord2018representation}, and SMILE \citep{song2020understanding}. We defer to \citet{pmlr-v97-poole19a} and \citet{guo_tight_2021} for summaries. Unfortunately, these variational methods do not directly apply to our problem setting. They involve modeling either the joint distribution $p(X,Y)$ or the generative distribution $p(X|Y)$. However, we consider the classification tasks where the state-of-the-art methods finetune the pretrained deep networks to model the conditional distributions of classification tasks $p(Y|X)$. It is possible to model the generative distribution on text classification datasets, but we consider that out of the scope of this paper. A recent paper, \citet{mcallester2020formal}, argues in favor of using (and minimizing) the difference of entropies to estimate the terms related to mutual information because, unlike DV, NWJ, MINE, and CPC, this setting is not restricted to various statistical limitations.

How about directly estimating the entropy values $H(Y\,|\,X)$ and $H(Y\,|\,X_s)$ from data? It turns out that the computational effort required by this approach can easily grow prohibitive. 
Estimating the conditional entropy from the dataset $\{x_i, y_i\}_{i=1..N}$ involves finding the density, which is usually implemented by finding the nearest neighbors. This could require $\mathcal{O}(N\text{log}N)$ computational time with $\mathcal{O}(N)$ memory\footnote{Store all data points using a heap-like data structure, which allows query in $\mathcal{O}(\text{log}N)$ time for each data point.} -- where the memory requirements would grow prohibitively -- or $\mathcal{O}(N^2)$ computational time with $\mathcal{O}(1)$ memory\footnote{Traverse the dataset to find nearest neighbors.} -- where the computational time would grow prohibitively. In comparison, training two models with stochastic gradient descent requires only $\mathcal{O}(N)$ training time and $\mathcal{O}(1)$ memory. In other words, our method is more realistic under real-world computational constraints.

We run Monte Carlo simulations on a fraction of data using an off-the-shelf entropy estimator, NPEET \citep{kraskov2004estimating}. The sizes of the fractions are decided to be stable following the analysis of \S \ref{subsection:experiments:dataset-size}, i.e., $10^3$ for IMDB and Yelp, $10^4$ for Quora, and $10^5$ for MNLI. We sample the subsets in a stratified manner with ten different random seeds. The conditional entropies $H(Y|X)$ and $H(Y|X_s)$ from Monte Carlo simulations differ significantly from those cross-entropy losses. Moreover, these simulations sometimes report negative $I(Y\,|\,X_t)$ values, indicating the prohibitive levels of the errors. We include the details in the Supplementary Data.

%we estimate $H(Y\,|\,X)$ and $H(Y\,|\,X_s)$ using data subsets, which can be unstable, dependent on how we vectorize $X$. For example, on MNLI with 1,000 samples, RoBERTa, BERT, and ALBERT vectorization would lead to $H(Y\,|\,X)$ of 97.14, 38.34, and 113.18 nats, respectively. This shows that directly sampling with a nonparametric approach is unstable.

\begin{table}[t]
\begin{center}
\begin{tabular}{c c c c c}
     \toprule
         \textbf{Dataset} &
         \textbf{$\text{Acc}_{Y\,|\,X}$} &
         \textbf{TSI$^{\text{P+S}}$} &
         \textbf{TSI$^{\text{P+S+O}}$}\\
         \midrule
            MNLI & 0.85 & 0.68 & 0.64 \\
        \midrule
            IMDB & 0.92 & 0.43 & -- \\
        \midrule
            Yelp & 0.97 & 0.41 & -- \\
        \midrule
            QQP & 0.89 & 0.31 & 0.23 \\
        \midrule
    \bottomrule
\end{tabular}
\end{center}
\caption{Our best estimates of TSI with P+S and P+S+O shortcut features respectively, and the dev accuracies of the corresponding $Y|X$ classifications.  
\label{tab:perf}}
\end{table}

\subsection{TSI required to classify each dataset}
\label{subsec:full-dataset}
Table \ref{tab:perf} contains our best estimations for TSI across datasets. The TSI$^{P+S}$ of IMDB and Yelp are similar. Moreover, both TSI$^{P+S}$ and TSI$^{P+S+O}$ of MNLI are about 0.4 nats larger than those of QQP. Considering that the highest dev accuracy on MNLI and QQP are similar, the contrast in TSI provides an alternative perspective in comparing across tasks. On the QQP dataset, neural models rely more on the artifacts, including punctuations and stopwords, than on the MNLI dataset.

Our method does not directly apply to HANS \citep{ThomasMcCoy2019} yet, since existing high-performing models mostly use HANS as a test set. Instead of directly approximating the TSI of HANS, one can compute that of, e.g., HANS + MNLI.

\subsection{Broader impacts}
While there is a general momentum to develop better models on miscellaneous classification tasks, we call for more systematic comparisons across different datasets and propose developing datasets with higher ``signal-to-noise ratios'', as measured by, e.g., TSI. We also encourage the NLP community to think about several closely related problems:

\textit{Identifying shortcut features.} While the release of a new NLP dataset is often paired with strong baselines for the proposed task, we also encourage future researchers to identify potential shortcuts or spurious associations, which could occur either due to the data collection procedure or due to the nature of the task itself (e.g., as reported by ~\citet{romanov2018lessons} for natural language inference tasks). 

\textit{Leaderboard practices.} Currently, the leaderboard practices reward high classification performances. We recommend that NLP researchers build leaderboards that additionally incentivize the minimal use of shortcuts. A potential way to do this would be constructing multiple test sets~\cite{glockner-etal-2018-breaking}, testing for different parameters of concern -- such as data efficiency, fairness, etc., -- as identified by ~\citet{ethayarajh-jurafsky-2020-utility}.

\textit{Metrics for cross-task comparison.} Consider reporting the performance on a unified scale of ``task-specific informativeness", rather than relying on average model performance metrics~\cite{Collins2018}. Designing metrics with grounds in linguistic knowledge is an interesting direction of future work. 

\section{Conclusion}
We propose a framework to quantify the task-specific information (TSI) for classifications on text-based datasets. Given a fixed collection of shortcut features, TSI quantifies the linguistic knowledge attributable to the classification target that is \textit{independent} of the shortcut features. The quantification method is computable under limited resources and is relatively robust to the dataset sizes. Further, this framework allows comparison across classification tasks under a standardized setting. For example, apart from the effects of punctuations and the non-negation stopwords, MNLI involves around 2.2 times TSI as the Quora Question Pairs, in terms of nats per sample.

%\citet{Damour2020,arjovsky2019invariant,Geirhos2020shortcut} provide a general discussion regarding underspecification and shortcut learning in machine learning, specifically as they relate to poor performance in the real world.

\section*{Acknowledgements}
We would like to thank the anonymous reviewers at NAACL 2021 and ACL (2022 August) for the feedback. Rudzicz is supported by a CIFAR Chair in artificial intelligence.

\bibliography{main}

\begin{thebibliography}{68}
\expandafter\ifx\csname natexlab\endcsname\relax\def\natexlab#1{#1}\fi

\bibitem[{Alain and Bengio(2017)}]{Alain2017}
Guillaume Alain and Yoshua Bengio. 2017.
\newblock \href {http://arxiv.org/abs/1610.01644} {{Understanding intermediate
  layers using linear classifier probes}}.
\newblock In \emph{ICLR}, Toulon, France.

\bibitem[{Barber and Agakov(2004)}]{agakov_im_2004}
David Barber and Felix Agakov. 2004.
\newblock \href
  {https://papers.nips.cc/paper/2003/file/a6ea8471c120fe8cc35a2954c9b9c595-Paper.pdf}
  {The {IM} algorithm: a variational approach to information maximization}.
\newblock \emph{Advances in neural information processing systems},
  16(320):201.

\bibitem[{Belghazi et~al.(2018)Belghazi, Baratin, Rajeshwar, Ozair, Bengio,
  Courville, and Hjelm}]{belghazi2018mutual}
Mohamed~Ishmael Belghazi, Aristide Baratin, Sai Rajeshwar, Sherjil Ozair,
  Yoshua Bengio, Aaron Courville, and Devon Hjelm. 2018.
\newblock \href {http://proceedings.mlr.press/v80/belghazi18a.html} {Mutual
  information neural estimation}.
\newblock In \emph{International Conference on Machine Learning}, pages
  531--540. PMLR.

\bibitem[{Bender and Koller(2020)}]{Bender2020}
Emily~M. Bender and Alexander Koller. 2020.
\newblock \href {https://doi.org/10.18653/v1/2020.acl-main.463} {{Climbing
  towards NLU: On Meaning, Form, and Understanding in the Age of Data}}.
\newblock In \emph{ACL}, pages 5185--5198, Online. Association for
  Computational Linguistics (ACL).

\bibitem[{Blache and Rauzy(2011)}]{blache2011predicting}
Philippe Blache and St{\'e}phane Rauzy. 2011.
\newblock Predicting linguistic difficulty by means of a morpho-syntactic
  probabilistic model.
\newblock In \emph{Proceedings of PACLIC}, pages 160--167.

\bibitem[{Chen et~al.(2019)Chen, Chu, and Gimpel}]{Chen2019DiscoEval}
Mingda Chen, Zewei Chu, and Kevin Gimpel. 2019.
\newblock \href {https://doi.org/10.18653/v1/D19-1060} {{Evaluation Benchmarks
  and Learning Criteria for Discourse-Aware Sentence Representations}}.
\newblock In \emph{EMNLP}, pages 649--662, Stroudsburg, PA, USA. Association
  for Computational Linguistics.

\bibitem[{Collins et~al.(2018)Collins, Rozanov, and Zhang}]{Collins2018}
Edward Collins, Nikolai Rozanov, and Bingbing Zhang. 2018.
\newblock \href {https://doi.org/10.18653/v1/k18-1037} {{Evolutionary data
  measures: Understanding the difficulty of text classification tasks}}.
\newblock In \emph{CoNLL}, pages 380--391. Association for Computational
  Linguistics.

\bibitem[{D'Amour et~al.(2020)D'Amour, Heller, Moldovan, and
  et~al.}]{Damour2020}
Alexander D'Amour, Katherine Heller, Dan Moldovan, and et~al. 2020.
\newblock \href {http://arxiv.org/abs/2011.03395} {{Underspecification in
  Machine Learning Underspecification Presents Challenges for Credibility in
  Modern Machine Learning}}.
\newblock Technical report, Google.

\bibitem[{Devlin et~al.(2019)Devlin, Chang, Lee, and
  Toutanova}]{devlin2019bert}
Jacob Devlin, Ming-Wei Chang, Kenton Lee, and Kristina Toutanova. 2019.
\newblock \href {https://www.aclweb.org/anthology/N19-1423/} {{BERT}:
  Pre-training of deep bidirectional transformers for language understanding}.
\newblock In \emph{NAACL}, pages 4171--4186.

\bibitem[{Donsker and Varadhan(1975)}]{donsker1975asymptotic}
Monroe~D Donsker and SR~Srinivasa Varadhan. 1975.
\newblock \href
  {https://onlinelibrary.wiley.com/doi/abs/10.1002/cpa.3160360204} {Asymptotic
  evaluation of certain markov process expectations for large time, i}.
\newblock \emph{Communications on Pure and Applied Mathematics}, 28(1):1--47.

\bibitem[{Doran et~al.(2018)Doran, Schulz, and Besold}]{doran2018does}
D~Doran, SC~Schulz, and TR~Besold. 2018.
\newblock \href {https://arxiv.org/abs/1710.00794} {What does explainable ai
  really mean? a new conceptualization of perspectives}.
\newblock In \emph{CEUR Workshop Proceedings}, volume 2071.

\bibitem[{Elazar et~al.(2021)Elazar, Ravfogel, Jacovi, and
  Goldberg}]{elazar_amnesic_2021}
Yanai Elazar, Shauli Ravfogel, Alon Jacovi, and Yoav Goldberg. 2021.
\newblock \href {http://arxiv.org/abs/2006.00995} {Amnesic {Probing}:
  {Behavioral} {Explanation} with {Amnesic} {Counterfactuals}}.
\newblock \emph{TACL}.
\newblock ArXiv: 2006.00995.

\bibitem[{Ethayarajh et~al.(2021)Ethayarajh, Choi, and
  Swayamdipta}]{ethayarajh-2021-information}
Kawin Ethayarajh, Yejin Choi, and Swabha Swayamdipta. 2021.
\newblock \href {https://kawine.github.io/assets/dataset_difficulty.pdf}
  {{Information-Theoretic Measures of Dataset Difficulty}}.

\bibitem[{Ethayarajh and Jurafsky(2020)}]{ethayarajh-jurafsky-2020-utility}
Kawin Ethayarajh and Dan Jurafsky. 2020.
\newblock \href {https://www.aclweb.org/anthology/2020.emnlp-main.393} {Utility
  is in the eye of the user: A critique of {NLP} leaderboards}.
\newblock In \emph{EMNLP}, pages 4846--4853, Online. Association for
  Computational Linguistics.

\bibitem[{Ettinger(2020)}]{ettinger2020bert}
Allyson Ettinger. 2020.
\newblock \href
  {https://www.mitpressjournals.org/doi/full/10.1162/tacl{\_}a{\_}00298} {{What
  BERT is not: Lessons from a new suite of psycholinguistic diagnostics for
  language models}}.
\newblock \emph{TACL}, 8:34--48.

\bibitem[{Gardner et~al.(2020)Gardner, Artzi, Basmova, Berant, Bogin, Chen,
  Dasigi, Dua, Elazar, Gottumukkala et~al.}]{gardner2020evaluating}
Matt Gardner, Yoav Artzi, Victoria Basmova, Jonathan Berant, Ben Bogin, Sihao
  Chen, Pradeep Dasigi, Dheeru Dua, Yanai Elazar, Ananth Gottumukkala, et~al.
  2020.
\newblock \href {https://arxiv.org/abs/2004.02709} {{Evaluating NLP Models via
  Contrast Sets}}.
\newblock \emph{arXiv preprint arXiv:2004.02709}.

\bibitem[{{Geirhos} et~al.(2020){Geirhos}, {Jacobsen}, {Michaelis}, {Zemel},
  {Brendel}, {Bethge}, and {Wichmann}}]{Geirhos2020shortcut}
Robert {Geirhos}, J{\"o}rn-Henrik {Jacobsen}, Claudio {Michaelis}, Richard
  {Zemel}, Wieland {Brendel}, Matthias {Bethge}, and Felix~A. {Wichmann}. 2020.
\newblock \href {https://arxiv.org/abs/2004.07780} {{Shortcut Learning in Deep
  Neural Networks}}.
\newblock \emph{arXiv e-prints}.

\bibitem[{Geirhos et~al.(2020)Geirhos, Jacobsen, Michaelis, Zemel, Brendel,
  Bethge, and Wichmann}]{geirhos2020unintended}
Robert Geirhos, J{\"o}rn-Henrik Jacobsen, Claudio Michaelis, Richard Zemel,
  Wieland Brendel, Matthias Bethge, and Felix~A Wichmann. 2020.
\newblock \href {https://jov.arvojournals.org/article.aspx?articleid=2771702}
  {Unintended cue learning: Lessons for deep learning from experimental
  psychology}.
\newblock \emph{Journal of Vision}, 20(11):652--652.

\bibitem[{Glockner et~al.(2018)Glockner, Shwartz, and
  Goldberg}]{glockner-etal-2018-breaking}
Max Glockner, Vered Shwartz, and Yoav Goldberg. 2018.
\newblock \href {https://doi.org/10.18653/v1/P18-2103} {Breaking {NLI} systems
  with sentences that require simple lexical inferences}.
\newblock In \emph{ACL}, pages 650--655, Melbourne, Australia. Association for
  Computational Linguistics.

\bibitem[{Guo et~al.(2021)Guo, Chen, Wang, Yang, Deng, Carin, Li, and
  Tao}]{guo_tight_2021}
Qing Guo, Junya Chen, Dong Wang, Yuewei Yang, Xinwei Deng, Lawrence Carin, Fan
  Li, and Chenyang Tao. 2021.
\newblock \href {http://arxiv.org/abs/2107.01131} {Tight {Mutual} {Information}
  {Estimation} {With} {Contrastive} {Fenchel}-{Legendre} {Optimization}}.
\newblock \emph{arXiv:2107.01131 [cs, math, stat]}.
\newblock ArXiv: 2107.01131.

\bibitem[{Gupta et~al.(2014)Gupta, Bengio, and Weston}]{Gupta2014}
Maya~R Gupta, Samy Bengio, and Jason Weston. 2014.
\newblock \href {https://jmlr.org/papers/volume15/gupta14a/gupta14a.pdf}
  {{Training Highly Multiclass Classifiers}}.
\newblock \emph{Journal of Machine Learning Research}, 15:1461--1492.

\bibitem[{Han et~al.(2020)Han, Wallace, and
  Tsvetkov}]{han-etal-2020-explaining}
Xiaochuang Han, Byron~C. Wallace, and Yulia Tsvetkov. 2020.
\newblock \href {https://doi.org/10.18653/v1/2020.acl-main.492} {Explaining
  black box predictions and unveiling data artifacts through influence
  functions}.
\newblock In \emph{ACL}, pages 5553--5563, Online. Association for
  Computational Linguistics.

\bibitem[{He et~al.(2019)He, Zha, and Wang}]{He2019}
He~He, Sheng Zha, and Haohan Wang. 2019.
\newblock \href {https://doi.org/10.18653/v1/d19-6115} {{Unlearn Dataset Bias
  in Natural Language Inference by Fitting the Residual}}.
\newblock In \emph{Workshop on Deep Learning Approaches for Low-Resource NLP},
  pages 132--142. Association for Computational Linguistics.

\bibitem[{Hewitt and Manning(2019)}]{hewitt-manning-2019-structural-probe}
John Hewitt and Christopher~D Manning. 2019.
\newblock \href {https://doi.org/10.18653/v1/N19-1419} {{A Structural Probe for
  Finding Syntax in Word Representations}}.
\newblock In \emph{NAACL}, pages 4129--4138, Minneapolis, Minnesota.
  Association for Computational Linguistics.

\bibitem[{Jain et~al.(2020)Jain, Patel, Nagalapatti, Gupta, Mehta, Guttula,
  Mujumdar, Afzal, {Sharma Mittal}, and Munigala}]{Jain2020DataQuality}
Abhinav Jain, Hima Patel, Lokesh Nagalapatti, Nitin Gupta, Sameep Mehta,
  Shanmukha Guttula, Shashank Mujumdar, Shazia Afzal, Ruhi {Sharma Mittal}, and
  Vitobha Munigala. 2020.
\newblock \href {https://doi.org/10.1145/3394486.3406477} {\emph{{Overview and
  Importance of Data Quality for Machine Learning Tasks}}}, pages 3561--3562.
  Association for Computing Machinery, New York, NY, USA.

\bibitem[{Jiang and de~Marneffe(2019)}]{jiang2019evaluating}
Nanjiang Jiang and Marie-Catherine de~Marneffe. 2019.
\newblock \href {https://www.aclweb.org/anthology/D19-1630} {{Evaluating BERT
  for natural language inference: A case study on the CommitmentBank}}.
\newblock In \emph{EMNLP-IJCNLP}, pages 6088--6093. Association for
  Computational Linguistics.

\bibitem[{Kaushik et~al.(2020)Kaushik, Hovy, and Lipton}]{kaushik2019learning}
Divyansh Kaushik, Eduard Hovy, and Zachary~C Lipton. 2020.
\newblock \href {http://arxiv.org/abs/1909.12434} {{Learning the difference
  that makes a difference with counterfactually-augmented data}}.
\newblock \emph{ICLR}.

\bibitem[{Kingma and Ba(2014)}]{kingma2014adam}
Diederik~P Kingma and Jimmy Ba. 2014.
\newblock Adam: A method for stochastic optimization.
\newblock \emph{arXiv preprint arXiv:1412.6980}.

\bibitem[{Koto et~al.(2021)Koto, Lau, and Baldwin}]{koto-etal-2021-discourse}
Fajri Koto, Jey~Han Lau, and Timothy Baldwin. 2021.
\newblock \href {https://doi.org/10.18653/v1/2021.naacl-main.301} {Discourse
  probing of pretrained language models}.
\newblock In \emph{NAACL}, pages 3849--3864, Online. Association for
  Computational Linguistics.

\bibitem[{Kraskov et~al.(2004)Kraskov, St{\"o}gbauer, and
  Grassberger}]{kraskov2004estimating}
Alexander Kraskov, Harald St{\"o}gbauer, and Peter Grassberger. 2004.
\newblock \href {https://journals.aps.org/pre/pdf/10.1103/PhysRevE.69.066138}
  {Estimating mutual information}.
\newblock \emph{Physical review E}, 69(6):066138.

\bibitem[{Kumar et~al.(2019)Kumar, Wintner, Smith, and
  Tsvetkov}]{kumar2019topics}
Sachin Kumar, Shuly Wintner, Noah~A Smith, and Yulia Tsvetkov. 2019.
\newblock \href {https://www.aclweb.org/anthology/D19-1425/} {Topics to avoid:
  Demoting latent confounds in text classification}.
\newblock In \emph{EMNLP-IJCNLP}, pages 4144--4154.

\bibitem[{Lakretz et~al.(2019)Lakretz, Kruszewski, Desbordes, Hupkes, Dehaene,
  and Baroni}]{lakretz-etal-2019-emergence}
Yair Lakretz, German Kruszewski, Theo Desbordes, Dieuwke Hupkes, Stanislas
  Dehaene, and Marco Baroni. 2019.
\newblock \href {https://doi.org/10.18653/v1/N19-1002} {{The emergence of
  number and syntax units in LSTM language models}}.
\newblock In \emph{NAACL}, pages 11--20, Minneapolis, Minnesota. Association
  for Computational Linguistics.

\bibitem[{Lan et~al.(2020)Lan, Chen, Goodman, Gimpel, Sharma, Soricut, and
  Research}]{Lan2020ALBERT}
Zhenzhong Lan, Mingda Chen, Sebastian Goodman, Kevin Gimpel, Piyush Sharma,
  Radu Soricut, and Google Research. 2020.
\newblock \href {http://arxiv.org/abs/1909.11942} {{ALBERT: A Lite BERT For
  Self-Supervised Learning of Language Representations}}.
\newblock In \emph{ICLR}.

\bibitem[{Li and Eisner(2019)}]{li-eisner-2019-specializing}
Xiang~Lisa Li and Jason Eisner. 2019.
\newblock \href {https://doi.org/10.18653/v1/D19-1276} {Specializing word
  embeddings (for parsing) by information bottleneck}.
\newblock In \emph{Proceedings of the 2019 Conference on Empirical Methods in
  Natural Language Processing and the 9th International Joint Conference on
  Natural Language Processing (EMNLP-IJCNLP)}, pages 2744--2754, Hong Kong,
  China. Association for Computational Linguistics.

\bibitem[{Liu et~al.(2019)Liu, Ott, Goyal, Du, Joshi, Chen, Levy, Lewis,
  Zettlemoyer, and Stoyanov}]{Liu2019roberta}
Yinhan Liu, Myle Ott, Naman Goyal, Jingfei Du, Mandar Joshi, Danqi Chen, Omer
  Levy, Mike Lewis, Luke Zettlemoyer, and Veselin Stoyanov. 2019.
\newblock \href {http://arxiv.org/abs/1907.11692} {{RoBERTa: A Robustly
  Optimized BERT Pretraining Approach}}.
\newblock \emph{arXiv:1907.11692}.

\bibitem[{Lovering et~al.(2021)Lovering, Jha, Linzen, and
  Pavlick}]{lovering_predicting_2021}
Charles Lovering, Rohan Jha, Tal Linzen, and Ellie Pavlick. 2021.
\newblock \href {https://openreview.net/forum?id=mNtmhaDkAr} {Predicting
  {Inductive} {Biases} of {Pre}-{Trained} {Models}}.
\newblock In \emph{{ICLR}}.

\bibitem[{Maas et~al.(2011)Maas, Daly, Pham, Huang, Ng, and
  Potts}]{Maas2011IMDB}
Andrew~L Maas, Raymond~E Daly, Peter~T Pham, Dan Huang, Andrew~Y Ng, and
  Christopher Potts. 2011.
\newblock \href {http://www.aclweb.org/anthology/P11-1015} {{Learning Word
  Vectors for Sentiment Analysis}}.
\newblock In \emph{ACL}, pages 142--150, Portland, Oregon, USA. Association for
  Computational Linguistics.

\bibitem[{McAllester and Stratos(2020)}]{mcallester2020formal}
David McAllester and Karl Stratos. 2020.
\newblock \href {http://proceedings.mlr.press/v108/mcallester20a.html} {Formal
  limitations on the measurement of mutual information}.
\newblock In \emph{International Conference on Artificial Intelligence and
  Statistics}, pages 875--884. PMLR.

\bibitem[{McCoy et~al.(2019)McCoy, Pavlick, and Linzen}]{ThomasMcCoy2019}
Tom McCoy, Ellie Pavlick, and Tal Linzen. 2019.
\newblock \href {https://doi.org/10.18653/v1/p19-1334} {{Right for the wrong
  reasons: Diagnosing syntactic heuristics in natural language inference}}.
\newblock In \emph{ACL}, pages 3428--3448. Association for Computational
  Linguistics.

\bibitem[{Misra et~al.(2020)Misra, Ettinger, and Rayz}]{Misra2020}
Kanishka Misra, Allyson Ettinger, and Julia~Taylor Rayz. 2020.
\newblock \href {http://arxiv.org/abs/2010.03010v1} {{Exploring BERT's
  Sensitivity to Lexical Cues using Tests from Semantic Priming}}.
\newblock In \emph{Findings of EMNLP}.

\bibitem[{Nguyen et~al.(2010)Nguyen, Wainwright, and
  Jordan}]{nguyen2010estimating}
XuanLong Nguyen, Martin~J Wainwright, and Michael~I Jordan. 2010.
\newblock \href {https://arxiv.org/abs/0809.0853} {Estimating divergence
  functionals and the likelihood ratio by convex risk minimization}.
\newblock \emph{IEEE Transactions on Information Theory}, 56(11):5847--5861.

\bibitem[{Niven and Kao(2020)}]{Niven2020}
Timothy Niven and Hung~Yu Kao. 2020.
\newblock \href {https://doi.org/10.18653/v1/p19-1459} {{Probing neural network
  comprehension of natural language arguments}}.
\newblock In \emph{ACL}, pages 4658--4664. Association for Computational
  Linguistics (ACL).

\bibitem[{O'Connor and Andreas(2021)}]{o2021context}
Joe O'Connor and Jacob Andreas. 2021.
\newblock \href {https://arxiv.org/abs/2106.08367} {What context features can
  transformer language models use?}
\newblock In \emph{ACL}. Association for Computational Linguistics.

\bibitem[{Oord et~al.(2018)Oord, Li, and Vinyals}]{oord2018representation}
Aaron van~den Oord, Yazhe Li, and Oriol Vinyals. 2018.
\newblock \href {https://arxiv.org/abs/1807.03748} {Representation learning
  with contrastive predictive coding}.
\newblock \emph{arXiv preprint arXiv:1807.03748}.

\bibitem[{Pedregosa et~al.(2011)Pedregosa, Varoquaux, Gramfort, Michel,
  Thirion, Grisel, Blondel, Prettenhofer, Weiss, Dubourg, Vanderplas, Passos,
  Cournapeau, Brucher, Perrot, and Duchesnay}]{scikit-learn}
F~Pedregosa, G~Varoquaux, A~Gramfort, V~Michel, B~Thirion, O~Grisel, M~Blondel,
  P~Prettenhofer, R~Weiss, V~Dubourg, J~Vanderplas, A~Passos, D~Cournapeau,
  M~Brucher, M~Perrot, and E~Duchesnay. 2011.
\newblock \href {https://scikit-learn.org/} {{Scikit-learn: Machine Learning in
  Python}}.
\newblock \emph{Journal of Machine Learning Research}, 12:2825--2830.

\bibitem[{Pimentel and Cotterell(2021)}]{pimentel2021bayesian}
Tiago Pimentel and Ryan Cotterell. 2021.
\newblock A bayesian framework for information-theoretic probing.
\newblock \emph{arXiv preprint arXiv:2109.03853}.

\bibitem[{Pimentel et~al.(2020)Pimentel, Valvoda, Maudslay, Zmigrod, Williams,
  and Cotterell}]{pimentel2020information}
Tiago Pimentel, Josef Valvoda, Rowan~Hall Maudslay, Ran Zmigrod, Adina
  Williams, and Ryan Cotterell. 2020.
\newblock \href {https://arxiv.org/abs/2004.03061} {{Information-Theoretic
  Probing for Linguistic Structure}}.
\newblock In \emph{ACL}. Association of Computational Linguistics.

\bibitem[{Poole et~al.(2019)Poole, Ozair, Van Den~Oord, Alemi, and
  Tucker}]{pmlr-v97-poole19a}
Ben Poole, Sherjil Ozair, Aaron Van Den~Oord, Alex Alemi, and George Tucker.
  2019.
\newblock \href {http://proceedings.mlr.press/v97/poole19a.html} {{On
  Variational Bounds of Mutual Information}}.
\newblock In \emph{ICML}, volume~97 of \emph{Proceedings of Machine Learning
  Research}, pages 5171--5180. PMLR.

\bibitem[{Pryzant et~al.(2021)Pryzant, Card, Jurafsky, Veitch, and
  Sridhar}]{pryzant-etal-2021-causal}
Reid Pryzant, Dallas Card, Dan Jurafsky, Victor Veitch, and Dhanya Sridhar.
  2021.
\newblock \href {https://doi.org/10.18653/v1/2021.naacl-main.323} {Causal
  effects of linguistic properties}.
\newblock In \emph{Proceedings of the 2021 Conference of the North American
  Chapter of the Association for Computational Linguistics: Human Language
  Technologies}, pages 4095--4109, Online. Association for Computational
  Linguistics.

\bibitem[{Romanov and Shivade(2018)}]{romanov2018lessons}
Alexey Romanov and Chaitanya Shivade. 2018.
\newblock \href {https://www.aclweb.org/anthology/D18-1187/} {Lessons from
  natural language inference in the clinical domain}.
\newblock In \emph{EMNLP}, pages 1586--1596. Association for Computational
  Linguistics.

\bibitem[{Rosenman et~al.(2020)Rosenman, Jacovi, and Goldberg}]{Rosenman2020}
Shachar Rosenman, Alon Jacovi, and Yoav Goldberg. 2020.
\newblock \href {http://arxiv.org/abs/2010.03656} {{Exposing Shallow Heuristics
  of Relation Extraction Models with Challenge Data}}.
\newblock In \emph{EMNLP}.

\bibitem[{Song and Ermon(2020)}]{song2020understanding}
Jiaming Song and Stefano Ermon. 2020.
\newblock \href {https://openreview.net/forum?id=B1x62TNtDS} {{Understanding
  the Limitations of Variational Mutual Information Estimators}}.
\newblock \emph{ICLR}.

\bibitem[{Stali and Iacobacci(2020)}]{Stali2020}
Ieva Stali and Ignacio Iacobacci. 2020.
\newblock \href {http://arxiv.org/abs/2009.08257} {{Compositional and Lexical
  Semantics in RoBERTa, BERT and DistilBERT: A Case Study on CoQA}}.
\newblock In \emph{EMNLP}.

\bibitem[{Steinke and Zakynthinou(2020)}]{steinke20ConditionalMI}
Thomas Steinke and Lydia Zakynthinou. 2020.
\newblock \href {http://proceedings.mlr.press/v125/steinke20a.html} {{Reasoning
  About Generalization via Conditional Mutual Information}}.
\newblock In \emph{COLT}, volume 125 of \emph{Proceedings of Machine Learning
  Research}, pages 3437--3452. PMLR.

\bibitem[{Swayamdipta et~al.(2020)Swayamdipta, Schwartz, Lourie, Wang,
  Hajishirzi, Smith, Choi, and Allen}]{Swayamdipta2020}
Swabha Swayamdipta, Roy Schwartz, Nicholas Lourie, Yizhong Wang, Hannaneh
  Hajishirzi, Noah~A Smith, Yejin Choi, and Allen. 2020.
\newblock \href {http://arxiv.org/abs/2009.10795} {{Dataset Cartography:
  Mapping and Diagnosing Datasets with Training Dynamics}}.
\newblock In \emph{EMNLP}.

\bibitem[{Tenney et~al.(2019)Tenney, Das, and Pavlick}]{tenney-etal-2019-bert}
Ian Tenney, Dipanjan Das, and Ellie Pavlick. 2019.
\newblock \href {https://doi.org/10.18653/v1/P19-1452} {{BERT Rediscovers the
  Classical NLP Pipeline}}.
\newblock In \emph{ACL}, pages 4593--4601, Florence, Italy. Association for
  Computational Linguistics.

\bibitem[{Tu et~al.(2020)Tu, Lalwani, Gella, and He}]{tu2020empirical}
Lifu Tu, Garima Lalwani, Spandana Gella, and He~He. 2020.
\newblock \href {https://arxiv.org/abs/2007.06778} {An empirical study on
  robustness to spurious correlations using pre-trained language models}.
\newblock \emph{TACL}.

\bibitem[{Voita and Titov(2020)}]{voita-titov-2020-information}
Elena Voita and Ivan Titov. 2020.
\newblock \href {https://doi.org/10.18653/v1/2020.emnlp-main.14}
  {Information-theoretic probing with minimum description length}.
\newblock In \emph{Proceedings of the 2020 Conference on Empirical Methods in
  Natural Language Processing (EMNLP)}, pages 183--196, Online. Association for
  Computational Linguistics.

\bibitem[{Wang et~al.(2019)Wang, Singh, Michael, Hill, Levy, and
  Bowman}]{Wang2019}
Alex Wang, Amanpreet Singh, Julian Michael, Felix Hill, Omer Levy, and Samuel~R
  Bowman. 2019.
\newblock \href {https://openreview.net/pdf?id=rJ4km2R5t7} {{GLUE: A Multi-Task
  Benchmark and Analysis Platform for Natural Language Understanding}}.
\newblock In \emph{ICLR}.

\bibitem[{Wang and Culotta(2020)}]{wang-culotta-2020-identifying}
Zhao Wang and Aron Culotta. 2020.
\newblock \href {https://www.aclweb.org/anthology/2020.findings-emnlp.308}
  {Identifying spurious correlations for robust text classification}.
\newblock In \emph{Findings of EMNLP}, pages 3431--3440, Online. Association
  for Computational Linguistics.

\bibitem[{Warstadt et~al.(2020)Warstadt, Zhang, Li, Liu, and
  Bowman}]{warstadt2020learning}
Alex Warstadt, Yian Zhang, Haau-Sing Li, Haokun Liu, and Samuel~R Bowman. 2020.
\newblock \href {https://www.aclweb.org/anthology/2020.emnlp-main.16/}
  {{Learning Which Features Matter: RoBERTa Acquires a Preference for
  Linguistic Generalizations (Eventually)}}.
\newblock \emph{EMNLP}.

\bibitem[{Williams et~al.(2018)Williams, Nangia, and Bowman}]{Multi-NLI}
Adina Williams, Nikita Nangia, and Samuel Bowman. 2018.
\newblock \href {http://aclweb.org/anthology/N18-1101} {{A Broad-Coverage
  Challenge Corpus for Sentence Understanding through Inference}}.
\newblock In \emph{NAACL}, pages 1112--1122. Association for Computational
  Linguistics.

\bibitem[{Wolf et~al.(2019)Wolf, Debut, Sanh, Chaumond, Delangue, Moi, Cistac,
  Rault, Louf, Funtowicz et~al.}]{wolf2019huggingface}
Thomas Wolf, Lysandre Debut, Victor Sanh, Julien Chaumond, Clement Delangue,
  Anthony Moi, Pierric Cistac, Tim Rault, R{\'e}mi Louf, Morgan Funtowicz,
  et~al. 2019.
\newblock \href {https://arxiv.org/abs/1910.03771} {Huggingface's transformers:
  State-of-the-art natural language processing}.
\newblock \emph{ArXiv}, pages arXiv--1910.

\bibitem[{Xu et~al.(2020)Xu, Zhao, Song, Stewart, and Ermon}]{xu2020theory}
Yilun Xu, Shengjia Zhao, Jiaming Song, Russell Stewart, and Stefano Ermon.
  2020.
\newblock \href {https://openreview.net/forum?id=r1eBeyHFDH} {A theory of
  usable information under computational constraints}.
\newblock \emph{ICLR}.

\bibitem[{Yang et~al.(2019)Yang, Dai, Yang, Carbonell, Salakhutdinov, and
  Le}]{yang2019xlnet}
Zhilin Yang, Zihang Dai, Yiming Yang, Jaime Carbonell, Russ~R Salakhutdinov,
  and Quoc~V Le. 2019.
\newblock \href {https://arxiv.org/abs/1906.08237} {{XLNet}: Generalized
  autoregressive pretraining for language understanding}.
\newblock In \emph{Advances in neural information processing systems}, pages
  5753--5763.

\bibitem[{Zhang et~al.(2015)Zhang, Zhao, and LeCun}]{zhang2015character_yelp}
Xiang Zhang, Junbo Zhao, and Yann LeCun. 2015.
\newblock \href
  {https://papers.nips.cc/paper/2015/file/250cf8b51c773f3f8dc8b4be867a9a02-Paper.pdf}
  {Character-level convolutional networks for text classification}.
\newblock In \emph{NIPS}, pages 649--657.

\bibitem[{Zhu et~al.(2020)Zhu, Pan, Abdalla, and Rudzicz}]{Zhu2020RSTprobe}
Zining Zhu, Chuer Pan, Mohamed Abdalla, and Frank Rudzicz. 2020.
\newblock \href {http://arxiv.org/abs/2010.00153} {{Examining the rhetorical
  capacities of neural language models}}.
\newblock In \emph{EMNLP BlackboxNLP Workshop}, pages 16--32. Association for
  Computational Linguistics.

\bibitem[{Zhu and Rudzicz(2020)}]{zhu-rudzicz-2020-information}
Zining Zhu and Frank Rudzicz. 2020.
\newblock \href {https://www.aclweb.org/anthology/2020.emnlp-main.744} {An
  information theoretic view on selecting linguistic probes}.
\newblock In \emph{EMNLP}, pages 9251--9262, Online. Association for
  Computational Linguistics.

\end{thebibliography}
\bibliographystyle{acl_natbib}

\newpage
\appendix

\section{Dataset details}
\label{sec:dataset-details}
\begin{itemize}[nosep]
    \item MNLI \citep{Multi-NLI} contains 392.7k English sentence pairs as train set. MNLI evaluates whether a model can detect entailment relationships between those pairs. They provided two dev sets: the ``matched'' and the ``mismatched'' portion. We take the ``matched'' portion (with 9.8k sentence pairs) as the dev set, since they are derived from the same sources as the sentences in the training set.
    \item IMDB \citep{Maas2011IMDB} is a large-scale dataset used to test a model's ability to detect sentiment from text. There are 50,000 movie reviews in English from IMDB in this dataset, with the training and dev sets containing 25,000 each.
    \item Yelp Reviews Polarity \citep{zhang2015character_yelp} contains 560k and 38k (in training and dev portion respectively) customer reviews in English from Yelp. These are collected to decide the polarity of opinions.
    \item Quora Question Pairs\footnote{\url{https://www.quora.com/q/quoradata/First-Quora-Dataset-Release-Question-Pairs}} contains 404k English question pairs on Quora, created to test the abilities of the models to understand the semantics from text, and determine whether the question pairs are synonymous. We randomly divide the train-dev-test data with 80-10-10 portions (with numpy random permutation, seed 0). 
\end{itemize}

\section{Hyperparameters}
\label{sec:hyperparameters}
Following list the search space of our hyperparameters for modeling $Y|X$.
\begin{itemize}[nosep]
    \item Optimizer: We use Adam optimizer \citep{kingma2014adam} to train the model parameters, and use the initial learning rate of lr$\in$\{2e-5, 1e-5\}.
    \item Train epochs: For full datasets, we run 3 epochs. For training subsets with $N\in \{10^5,10^4,10^3\}$ samples, we run either 3 or 10 epochs. For training the small $N=100$ sample subsets, we run $\{3, 10, 100\}$ epochs.
    \item Batch size: We run with batch sizes of $B\in \{2, 4, 8, 16\}$ for each classification setting. We find that in general, larger per-device batch sizes (e.g., 8 and 16) are better than smaller batches (e.g., 2 and 4), but a batch size of 16 or 32 could lead to out-of-memory issues on machines with 64GB memory.
\end{itemize}
Following the training procedure, our best development accuracies are comparable to the results reported on, e.g., the GLUE Benchmark leaderboard. While previous work added additional steps (e.g., learning rate warmup) to boost accuracy, our aim is not to beat the SOTA, but to establish a principled method that allows cross-task comparison. We include the hyperparameter configurations of all runs in the Supplementary Data.

For modelling $Y\,|\,X_s$, we use the scikit-learn \citep{scikit-learn} MLPClassifier with hidden sizes from \{10, 30, 100, 300, 10-10, 30-30, 100-100\} where, e.g., 10-10 indicates two hidden layers with 10 units each. We rely on the default training procedures, search for the optimal hidden sizes based on the validation losses, and report the dev loss $\text{NLL}(Y\,|\,X_s)$ scores.

\end{document}